\documentclass{article}

\PassOptionsToPackage{numbers}{natbib}
\usepackage[preprint]{neurips_2019}

\usepackage[utf8]{inputenc} 
\usepackage[T1]{fontenc}    
\usepackage{hyperref}       
\usepackage{url}            
\usepackage{booktabs}       
\usepackage{amsfonts}       
\usepackage{nicefrac}       
\usepackage{microtype}      

\usepackage[ruled,vlined]{algorithm2e} 

\usepackage{array}
\usepackage{enumitem}
\usepackage{amsthm} 
\usepackage[pdftex]{graphicx}
\usepackage{amsmath} 
\usepackage{amssymb} 
\usepackage{subfigure}
\usepackage{multicol}
\usepackage{wrapfig}

\title{Federated Forest}

\author{
Yang Liu$^{1,2}$\ , Yingting Liu$^{3,}$\thanks{Equal contribution. The research was done when the second and third authors were interns at JD Intelligent Cities Research.}\ , Zhijie Liu$^{4,}$\footnotemark[1]\ ,  Junbo Zhang$^{1,2,}$\thanks{Junbo Zhang is the corresponding author.}\ , Chuishi Meng$^{1,2}$\ , Yu Zheng$^{1,2}$\\
$^{1}$JD Intelligent Cities Research, Beijing, China\\
$^{2}$JD Intelligent Cities Business Unit, Beijing, China\\
$^{3}$University of Science and Technology of China, Hefei, China\\
$^{4}$Beijing Normal University, Beijing, China\\
\texttt{\{liuyang21cn,yingting6,msjunbozhang,msyuzheng\}@outlook.com}\\
\texttt{
201621210044@mail.bnu.edu.cn, chuishimeng@gmail.com}
}

\begin{document}
\maketitle

\begin{abstract}
Most real-world data are scattered across different companies or government organizations, and cannot be easily integrated under data privacy and related regulations such as the European Union's General Data Protection Regulation (GDPR) and China' Cyber Security Law. Such \textit{data islands} situation and \textit{data privacy \& security} are two major challenges for applications of artificial intelligence. In this paper, we tackle these challenges and propose a privacy-preserving machine learning model, called \textit{Federated Forest}, which is a lossless learning model of the traditional random forest method, i.e., achieving the same level of accuracy as the non-privacy-preserving approach. Based on it, we developed a secure cross-regional machine learning system that allows a learning process to be jointly trained over different regions' clients with the same user samples but different attribute sets, processing the data stored in each of them without exchanging their raw data. A novel prediction algorithm was also proposed which could largely reduce the communication overhead. 
Experiments on both real-world and UCI data sets demonstrate the performance of the Federated Forest is as accurate as the non-federated version.
The efficiency and robustness of our proposed system had been verified. Overall, our model is practical, scalable and extensible for real-life tasks.
  
\end{abstract}

\section{Introduction}
Artificial intelligence has made great progress in recent years thanks to the large amount of data collected in different domains. Unfortunately, the data has also arisen to be the largest bottleneck for the implementation of AI methods.
In real-world applications, the big data are scattered across different companies or government organizations and stored in the form of \textit{data islands}, in other words, 
data across different domains cannot be shared with each other. 
For companies the data is among one of the most important assets of companies which cannot be easily shared. 
Governments' data are highly secured and mostly not utilized.
Besides, people now are highly sensitive about data privacy. Data breaches happen occasionally and most countries now either have data privacy-related legislation enacted or being drafted. In 2018, the European Union enacted the General Data Protection Regulation (GDPR) \citep{regulation2016regulation}. 
The GDPR provides individuals with more control over their personal data and states strict principles and absolute transparencies on how businesses should handle these data.
Any type of tracking or record of personal data must be authorized by the customer before collection and business must clearly state their intentions and plans for the data. Faced with the difficulties and restrictions, the question becomes if it is worth investing effort to make use of the scattered data.  

\begin{wrapfigure}[13]{r}{0.5\textwidth}
\vspace{-.3cm}
\small
    \includegraphics[width=\linewidth]{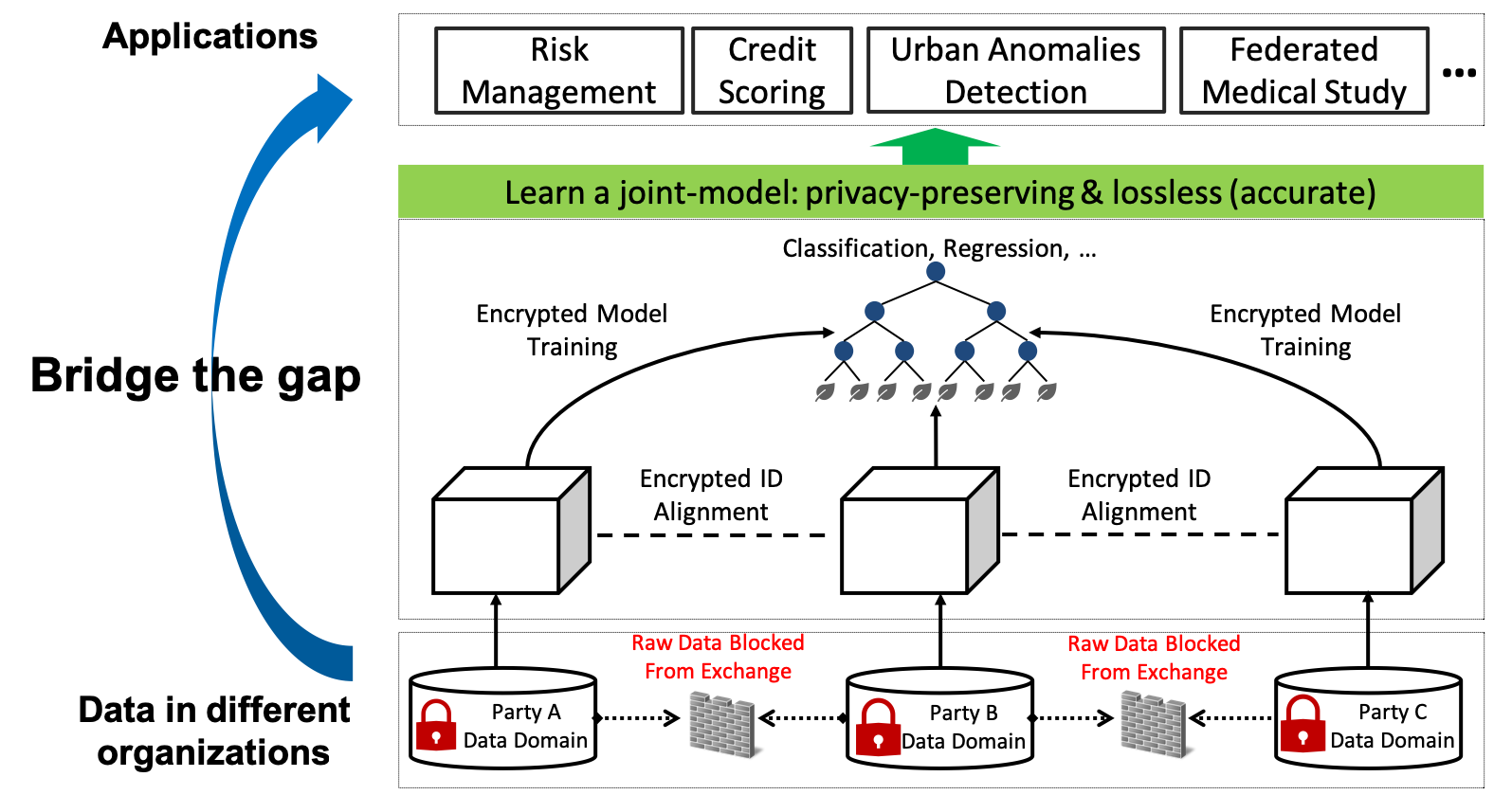}
    \caption{New Era of Machine Learning}
    \label{fig:intro}
\end{wrapfigure}

The answer is yes. Academia, companies and governments could all benefit from resolving the data islands situation. The joint-models are able to improve many current services and products, and support more potential applications, including but not limited to medical study, targeted marketing, urban anomalies detection and risk management, as shown in Figure \ref{fig:intro}. For example, banks could train joint-models with e-commerce companies to achieve a precised customer profiling and improve their marketing strategies. Government organizations could work with ride-hailing companies to have a better understanding of city's daily traffic flow and adjust the road planing based on it.

Consequently, the question becomes how can we train the joint-models. Faced with the challenges of \textit{data islands} and \textit{data privacy \& security}, the current available methods cannot completely solve the problems. Because of this, developing new methods to bridge the gap between real-world applications and data islands becomes an urgent problem. In 2016, a new approach named federated learning \citep{mcmahan2016communication, konevcny2016federated, OptimizationKone2016Federated} was proposed, which mainly focuses on building privacy-preserved machine learning models when data are distributed in different places. Federated learning has provided a new approach to look at the current problems, and shwon the possibility of real-life applications. 

Inspired by their work, we proposed a novel privacy-preserving tree-based machine learning model, named \textit{Federated Forest (FF)}. Based on it, we developed a secure cross-regional machine learning system, which is capable of conquering the challenges described above. Our contributions are four-folds: 
\begin{itemize}[leftmargin=*]
\item \textbf{\textit{Secured privacy.}} 
Data privacy is fully protected by redesigning the tree building algorithms, applying encryption methods and establishing a third-party trusty server. The contents and amount of information exchange are limited to a minimum, and each participant is blind to others.   
    
\item \textbf{\textit{Lossless (accurate).}} Our model is based on the methodologies of CART \citep{breiman2017classification} and bagging \citep{breiman1996bagging}, and fits the vertical federated setting. We experimentally proved that our model can achieve the same level of accuracy as the non-federated approach that brings the data into one place.
    
\item \textbf{\textit{Efficiency.}}
An efficient communication mechanism was implemented with methods of \textit{MPI} \citep{MPI} for sharing of the intermediate values. 
A fast prediction algorithm was designed and it's weakly correlated (scale-free) to the number of domains and trees, maximum tree depth and sample size.

\item \textbf{\textit{Practicability and scalability.}} Our model supports both classification and regression tasks, and is strongly practical, extensible and scalable for real-life applications. The experiments on real-world data sets had proved our model's accuracy, efficiency and robustness. 
\end{itemize}

\section{Related Work}
\subsection{Federated Learning}
Federated learning \citep{mcmahan2016communication, konevcny2016federated, OptimizationKone2016Federated} was first proposed to solve the problems that rich data are generated from user devices, but due to regulations it's difficult to build models from the data. The solution is to keep the data on user devices and train a shared model by aggregating locally calculated intermediate results in neural networks. In \citep{chen2018federated} they proposed a new recommender system which applies federated learning to meta-learning. Federated learning has also been applied to solve multi-task problems in \citep{smith2017federated} and a loss-based AdaBoost method was developed in \citep{huang2018loadaboost}. \citep{Hardy2017Private} introduced a vertically-aggregated federated learning method. In their work, each data provider possessed unique features, and sample IDs are aligned between them. They jointly learned a logistic regression model to secure the data privacy and keep modeling accurate. In addition, a modular benchmarking framework for federated settings was presented in the work of \citep{caldas2018leaf}. Although many research products have been coming out, the definition of federated learning was still blurry until the work of \citep{yang2019federated}. They categorized current federated learning methods into three types, horizontal federated learning, vertical federated learning and federated transfer learning. Following this survey, the same team introduced a new framework known as secure federated transfer learning \citep{liu2018secure} to build models for target-domain party by leveraging rich labels from source-domain party, as the data sets of the two parties are different in both sample space and feature space. In \citep{cheng2019secureboost} they reviewed the tree-boosting method and applied it to the vertical federated setting. A lossless framework was proposed and it was able to keep information of each private data provider from being revealed. In \citep{zhuo2019federated} they presented a novel reinforcement learning approach that considers the privacy requirement and builds Q-network for each agent with the help of other agents. To make the federated machine learning more practical, they are pushing to build a \hyperlink{https://www.fedai.org}{Federated AI Ecosystem} such that the partners can fully exploit their data's value and promote vertical applications. An IEEE standard \hyperlink{https://sagroups.ieee.org/3652-1/}{\textit{Guide For Architectural Framework And Application Of Federated Machine Learning}} \citep{FML} was also initialized and is being drafted. 

\subsection{Data Privacy Protection}
In federated learning, there are two major encryption methods applied for protecting data privacy and security, which are differential privacy \citep{dwork2011differential} and homomorphic encryption \citep{gentry2009fully}. The idea of differential privacy is to add properly calibrated noise to the algorithm or the data, with examples including \citep{Geyer2017, mcmahan2017learning}. This approach will not affect computational efficiency too much but may weaken model performance. Homomorphic encryption is a method that supports secure multiplication and addition on encrypted data, and once the result is decrypted, it should match the output of operations on the corresponding raw data. The work of \citep{hardy2017, Le2018Privacy, kim2018privacy} all used this approach. 
There are two major drawbacks of homomorphic encryption. First, the complexity of the algorithm is high and it will be intensely time consuming for frequent use. Second, it does not support operations of non-linear functions, such as Sigmoid and Logarithmic function, and approximations are necessary. In the work of \citep{Hardy2017Private} they used the Taylor expansion to approximate the Sigmoid function and \citep{kim2018secure} used least squares method. In theory these approaches could work but in our practice the results were not ideal. 

\section{Problem Formulation}

\subsection{Data Distribution}
In our work, we focus on the vertical federated learning problems, in which all participants have the same sample space but different feature space, as shown in Figure \ref{fig:federated-random-forest}. Consider each company or government organization as a regional data domain, denoted as $\mathcal{D}_{i}$, then the overall data domain is $\mathcal{D}  = \mathcal{D}_{1} \cup \mathcal{D}_{2} \cup \cdots \cup \mathcal{D}_{M}$, where $1\leq i \leq M$. $M$ is the number of regional domains. We denote the feature space of $\mathcal{D}_{i}$ as $\mathcal{F}_{i}$, then the entire feature space $\mathcal{F}$ is $\mathcal{F} = \mathcal{F}_{1} \cup \mathcal{F}_{2} \cup \cdots \cup \mathcal{F}_{M}$. During the modeling process, all features' true names were encoded to protect privacy. For any $i$ and $j$, if $i \neq j$ and $1 \leq i, j \leq M$, then $\mathcal{F}_{i} \cap \mathcal{F}_{j} = \varnothing$. In our work, all domains have the same number of samples and the sample IDs were aligned across domains. One master machine was deployed as the parameter server and multiple client machines were used, where each contains one regional data domain. The labels $y$ were provided by one of the clients, which we assume to be client $1$. Then the labels were copied to the master and clients in encrypted forms. Two things to notice here: 1) In reality, $M$ is usually small and even $M=5$ means there are five different organizations modeling together, which could be rare. The model design can be totally different for large $M$. 2) We are not going to talk about the methods of ID alignment since it is another research topic, discussed in work such as \citep{nock2018entity}. The notations appeared in this paper are also shown in Table \ref{Notations1}.


\subsection{Problem Statement}
The formal statement of the problem is given as below:

\textbf{Given:} Regional domain $\mathcal{D}_i$ and encrypted label $y$ on each client $i$, $1 \leq i \leq M$.

\textbf{Learn:} A Federated Forest, such that for each tree in the forest: 1) a complete tree model $T$ is held on master; 2) a partial tree model $T_i$ is stored on each client $i$, $1 \leq i \leq M$. 

\textbf{Constraint:} The performance (accuracy, f1-score, MSE, e.t.c.) of the Federated Forest must be comparable to the non-federated random forest.

\newpage

\section{Methodology}

\begin{wrapfigure}[11]{r}{0.5\textwidth}
\vspace{-1.3cm}
\small
    \includegraphics[width=\linewidth]{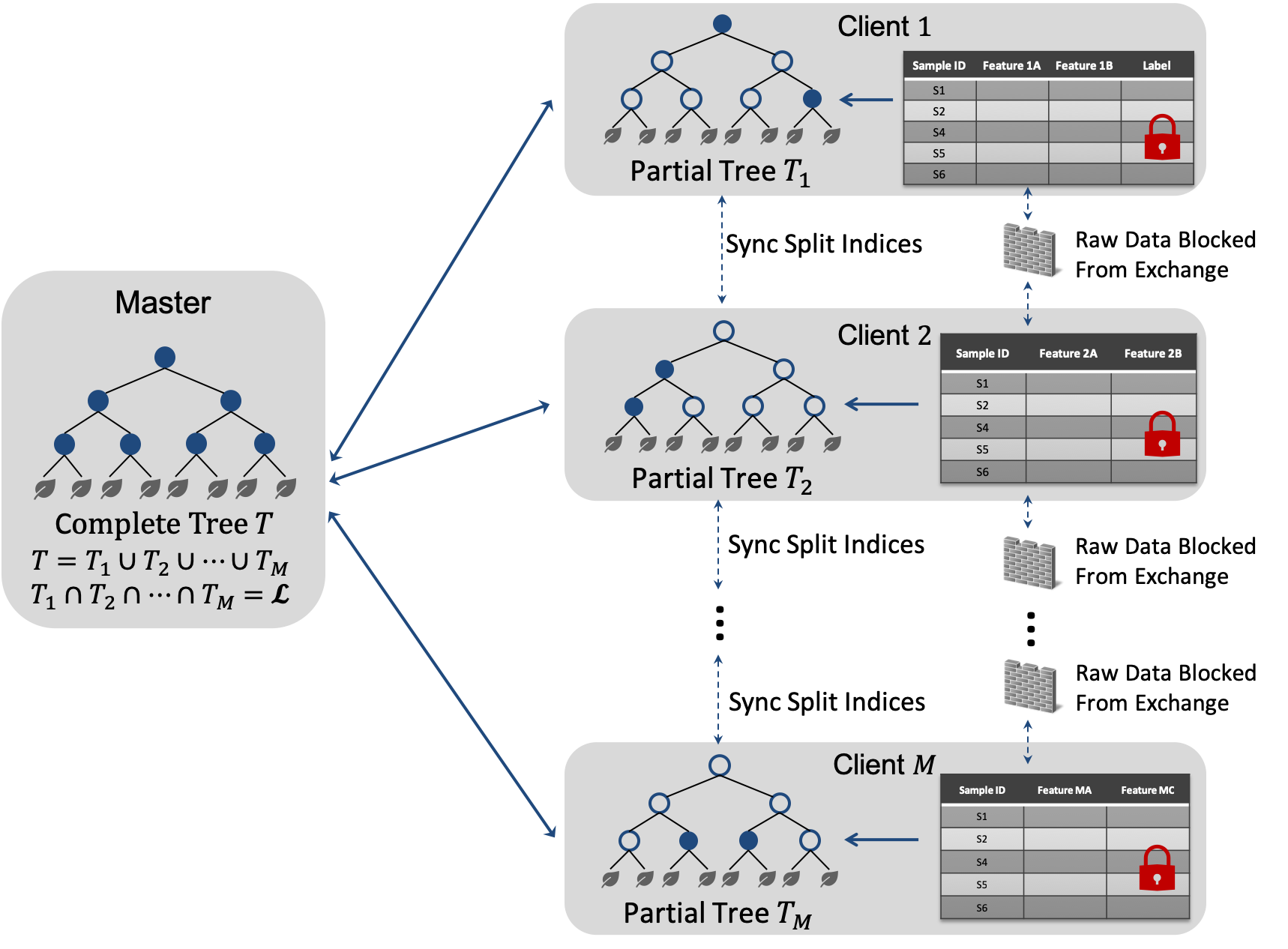}
    \caption{Federated Forest}
    \label{fig:federated-random-forest}
\end{wrapfigure}

Here we present the framework of Federated Forest, which is based on the CART tree \citep{breiman2017classification} and bagging \citep{breiman1996bagging}, and is able to deal with both classification and regression problems. The framework is shown as in Figure \ref{fig:federated-random-forest} and details of the algorithms are given in the following subsections.

\subsection{Model Building}

\paragraph{Algorithm.}
In our work, each tree is built by all parties working together and the tree structure is stored on the master node and every client. However, each tree only stores the split information with respect to their own features. We first present the client-side Federated Forest algorithm in Algorithm \ref{alg: dt-client}, and in Algorithm \ref{alg: dt-master} we described how the master coordinates the modeling process. 

\begin{wrapfigure}[35]{r}{0.56\textwidth}
\begin{minipage}{0.56\textwidth}
\vspace{-.4cm}
\begin{algorithm}[H]
    \small
    \SetKwInOut{Input}{Input}\SetKwInOut{Output}{Output}
    \SetKwFunction{Tree}{TreeBuild}%
	\Input{Data set $\mathcal{D}_{i}$ on client $i$;\\
	Local features $\mathcal F_{i} =\emptyset$ or $ \mathcal F_{i} = \{f_{A}, f_{B}, \cdots \}$;\\
	Encrypted label $y$;}
	\Output{Partial Federated Forest Model on Client $i$}
	\SetKwProg{TreeGenerateAlg}{Function}{}{end}
	\While{tree\_build is True}{
    Receive $\mathcal{F}_i^{'} \subset \mathcal{F}_i$ and $\mathcal{D}_i^{'} \subset \mathcal{D}_i$ for current tree building;\\
    \TreeGenerateAlg{\Tree($\mathcal{D}_{i}^{'}$, $\mathcal{F}_i^{'}$, $y$)}{
    Create empty tree node;\\
    \If{the pre-pruning condition is satisfied}{
    Mark current node as leaf node;\\
    Assign leaf label by voting;\\ 
    \Return{leaf node};}
    $p, f^{*} \leftarrow -\infty, None$;\\
	\If{$\mathcal F_{i}^{'} \neq\emptyset$}{
	Compute impurity improvement $p$ for any $f \in \mathcal F_{i}^{'}$ and find local maximum $p_i$;\\
	Record local best split feature $f^{*}$ and split threshold;\\}
	Send encrypted $p_i$ to master;\\
	\eIf{receive the split message from master}{
		/* Global best split feature is from itself */\\
		is\_selected $\leftarrow$ True; \\
		Split samples and send sample indices of left and right subtrees to master;}{Receive sample indices of left and right subtrees;\\}
    left\_subtree $\leftarrow$ \Tree($\mathcal D_{i\_left}^{'}$, $\mathcal{F}_i^{'}$, $y_{left}$);\\
	right\_subtree $\leftarrow$ \Tree($\mathcal D_{i\_right}^{'}$, $\mathcal{F}_i^{'}$, $y_{right}$);\\
	\If{is\_selected is True}{
	Save $f^{*}$ and split threshold to tree node;}
	Save subtrees to tree node;\\
	\Return{tree node;}}
	Append current tree to forest;\\}
	\Return{Partial Federated Forest Model on Client $i$;}
	\small{\caption{Federated Forest -- Client\label{alg: dt-client}}}
    \end{algorithm}
    \end{minipage}
\end{wrapfigure}

Following the bagging paradigm, the master node first randomly selects a subset of features and samples from the entire data. Then the master will notify each client the selected features and sample IDs privately. For the selected features, master will notify each client privately. For example, if ten features are chosen by the master and client 1 only possesses three of them, then client 1 will only know these three features were selected. It will never know how many features were chosen globally, not to mention what the features were. During the tree construction, the pre-pruning conditions are frequently checked. If the conditions are satisfied, the clients and master will create leaf nodes accordingly.

If the termination condition is not triggered, all clients enter the splitting state, and the best split feature of the current tree node will be selected by comparing the impurity improvements. First, each client $i$ finds the local optimal split feature $f_i^*$. Then the master collects all local optimal features and corresponding impurity improvements, allowing the global best feature to be found. Second, the master notifies the client who provided the global best feature. The corresponding client will split the samples and send the data partition results (sample IDs that fall into left and right subtrees) to the master for distribution. For the current tree node, only the client that provides the best split feature will save the details of this split. The other clients are only aware that the selected feature is not contributed by themselves. The split information such as threshold and split feature are also unknown to them. Last, the subtrees are recursively created and the current tree node is returned. In modeling, if the child trees nodes are created successfully, the parent node doesn't need to save the sample IDs for the subtrees. Otherwise, if the connection is down, the modeling can be easily recovered from the break point. 
\begin{wrapfigure}[24]{r}{0.56\textwidth}
\begin{minipage}{0.56\textwidth}
\begin{algorithm}[H]
    \small
	\SetKwInOut{Input}{Input}\SetKwInOut{Output}{Output}
	\SetKwFunction{Tree}{TreeBuild}%
	\SetKwProg{ForestGenerateAlg}{Function}{}{end}
	\Input{Indices of $\mathcal{D}$; \\
		Encoded features $\mathcal F = \mathcal{F}_{1} \cup \mathcal{F}_{2} \cup \cdots \cup \mathcal{F}_{M}$; \\
		Encrypted label $y$;\\}
	\Output{Complete Federated Forest Model}
		/*Build trees for forest recurrently*/\\
		\While{tree\_build is True}{
		Broadcast randomly selected samples $\mathcal D^{'}$;\\
        Randomly select features $\mathcal{F}_i^{'}$ from $\mathcal{F}_i$ and send to client $i$;\\			
			\ForestGenerateAlg{	\Tree($\mathcal D^{'}$, $\mathcal F^{'}$, $y$)}{

	            Create empty tree node;\\
				\If{the pre-pruning condition is satisfied}{
					    Mark current node as leaf node;\\
				 		Assign leaf label by voting;\\
						\Return{leaf node};}
				Receive encrypted $\{p\}_{i=1}^M$ and related information from all clients;\\
				Take $j = argmax(\{p\}_{i=1}^M)$ and notify client $j$;\\
				Receive split indices from client $j$ and broadcast;\\
				left\_subtree $\leftarrow$ \Tree($\mathcal D_{left}^{'}, \mathcal F^{'}, y_{left}$);\\
				right\_subtree $\leftarrow$ \Tree($\mathcal D_{right}^{'}, \mathcal F^{'}, y_{right}$);\\
				Save subtrees and split info to tree node;\\
				\Return{tree node;}
			}
			Append current tree to forest;\\
		}
		\Return{Complete Federated Forest Model};
		\small{\caption{Federated Forest -- Master	\label{alg: dt-master}}}
\end{algorithm}
\end{minipage}
\end{wrapfigure}

\paragraph{Model Storage.}
A tree predictive model is composed of two parts, tree structure and split information such as feature and threshold used for each split. Since the forest is built with all clients working together, the structure of each tree on every client is the same. However, for a given tree node, the client may or may not store the detailed information. Only the master server is able to store the complete model. For each tree node, the client will store the corresponding split threshold only if it provided the split feature. If not, the client will store nothing at the current node but only keep the node structure. We denoted the complete tree nodes as $T$, the one saved on master, and denoted the tree nodes without full details stored by $i$th client as $T_{i}$. Since the tree structure is consistent, we consider $T_{i} \subset T$, and $T_{1} \cap T_2 \cap \cdots \cap T_{M} = \mathcal{L}$, where $\mathcal{L}$ is the leaf node sets. The complete tree $T$ is the union of all partial trees, that $T = T_{1} \cup T_{2} \cup \cdots \cup T_{M}$. 



\subsection{Model Prediction}
\begin{wrapfigure}[23]{l}{0.56\textwidth}\begin{minipage}{0.56\textwidth}
\vspace{-.5cm}
\begin{algorithm}[H]
    \small
	\SetKwInOut{Input}{Input}\SetKwInOut{Output}{Output}
	\Input{Partial federated forest model saved on $i$th client;\\
		Encoded features $\mathcal F_{i}$ on $i$th client;\\
		Test set $\mathcal D^{test}_{i}$ on $i$th client;}
	\Output{Samples IDs $S_i^l$ of leaf $l$ on $T_i$, $l \in \mathcal{L}$}
	\SetKwProg{TreePredictAlg}{Function}{}{end}
	\SetKwFunction{Tree}{TreePredict}%
	\While{TreePrediction is True}{
	\TreePredictAlg{\Tree(\small{$T_{i}$, $\mathcal D^{test}_{i}$, $\mathcal F_{i}$})}{
		\eIf{is\_leaf is True}{Return sample IDs $S_{i}^{l}$ and leaf label;}{
			\eIf{$T_{i}$ keeps the split info of current node}
			{Split samples into subtrees;\\
				left\_subtree $\leftarrow$ \Tree(\small{{$T_{i\_left}$, $\mathcal F_{i}$, $\mathcal D_{i,left}^{test}$}});\\
				right\_subtree $\leftarrow$ \Tree(\small{{$T_{i\_right}$, $\mathcal F_{i}$, $\mathcal D_{i\_right}^{test}$}});\\
			}
			{
				left\_subtree $\leftarrow$ \Tree(\small{{$T_{i\_left}$, $\mathcal F_{i}$, $\mathcal D_{i}^{test}$}});\\
				right\_subtree $\leftarrow$ \Tree(\small{{$T_{i\_right}$, $\mathcal F_{i}$, $\mathcal D_{i}^{test}$}});\\
			}
			Return left and right subtrees;
		}	
		Send $S_{i} = \{S_{i}^{1}, S_{i}^{2}, \cdots ,S_{i}^{l}, \cdots \}$ to master;\\
	}}
	\caption{Federated Forest Prediction -- Client}
	\label{alg: pred-client}
\end{algorithm}
\end{minipage}
\end{wrapfigure}

Under the vertical federated setting \citep{yang2019federated}, the classical approach of prediction involves multiple rounds of communication between the master and clients, even for only one sample. When the number of trees, maximum tree depth and sample size are large, the communication requirements for predicting will become a serious burden. To address this problem, we designed a novel prediction method which takes the advantage of our distributed model storage strategy. Our method only needs one round of collective communication for each tree and even for the overall forest. We first present the prediction algorithm of the client side in Algorithm \ref{alg: pred-client}, and in Algorithm \ref{alg: pred-master}, we described how the master server coordinates each client to achieve the final predictions.

First, each client uses the locally stored model to predict samples. For the tree $T_i$ on $i$th client, each sample enters $T_{i}$ from the root node, and finally falls into one or several leaf nodes through the binary tree. When the sample travels through each node, if the model stores the split information at this node, then this sample is determined to enter the left or right subtree by checking the split threshold. If the model does not have split information at this node, the sample simultaneously enters both left and right subtrees. 

Secondly, the path determination of the tree node is performed recursively until each sample falls into one or several leaf nodes. When this process is finished, each leaf node of tree $T_i$ on client $i$ will keep a batch of samples. We use $S^{l}_{i}$ to represent the samples that fall into the leaf node $l$ of the tree model $T_{i}$, where $l \in \mathcal{L}$. $\mathcal{L}$ is the set of leaf nodes of the tree $T_{i}$. 

\begin{wrapfigure}[11]{r}{0.56\textwidth}\begin{minipage}{0.56\textwidth}
\vspace{-.5cm}
\begin{algorithm}[H]
    \small
	\SetKwInOut{Input}{Input}\SetKwInOut{Output}{Output}
	\Input{Sample IDs $S$ of test set $\mathcal{D}^{test}$\\}
	\Output{ Prediction of Federated Forest}
	\While{TreePrediction is True}{
		Gather $\{S_{1}, S_{2}, \cdots ,S_{i}, \cdots \}$;\\
		Obtain $\{ S^{1}, S^{2}, \cdots , S^{l}, \cdots \} $, where $S^{l} = S^{l}_{1} \cap S^{l}_{2} \cap \cdots \cap S^{l}_{M}$;\\
		Return label of leaf $l$ for samples in $S^l$, $l \in \mathcal{L}$;}
    Calculate forest predictions by voting on the results of trees;\\
    \Return{Final Predictions};
	\caption{Federated Forest Prediction -- Master}
	\label{alg: pred-master}
\end{algorithm}
\end{minipage}
\end{wrapfigure}

Thirdly, for each leaf $l\in \mathcal{L}$, the master will take the intersection on $\{S^{l}_{i}\}_{i=1}^{M}$, and the result will be $S^l$. Then the sample sets $S^l$ owned by each leaf node on complete tree $T$ are already associated with final predictions. Here we gave a formal proposition on our new prediction method so it can be mathematically defined:


\newtheorem{prop}{proposition}
\begin{prop}\label{prop}
For samples $S$ fall into one or multiple leaves on tree $T_i$, then for any leaf $l$ of the complete tree $T$, the sample IDs $S^{l}$ in leaf $l$ can be obtained by taking intersection of $\{S^{l}_{i}\}_{i=1}^{M}$, that $S^{l} = S^{l}_{1} \cap S^{l}_{2} \cap \cdots \cap S^{l}_{M}$.
\end{prop}
The proof is provided in Appendix \ref{Proof}. After obtaining the label values for each sample on all trees, we can easily achieve final predictions. In this approach, we only need one round of communication for each tree, or even only one round for the entire forest.

\subsection{Privacy Protection}

Here we have categorized our efforts on the privacy protection into five parts:

\textit{\textbf{Identities.}}
In real world tasks, we often face situations where IDs of samples are tied to persons' real identities. Because of this, we have to encrypt the identities before the ID alignment. An example approach could be like following: First all clients use an agreed hash method to transform the sample IDs and generate new hashed IDs. Then Message-Digest Algorithm 5 (MD5) can be applied on the hashed IDs and generate irreversibly encrypted IDs.

\textit {\textbf{Labels.}}
For classification problems, even labels are encoded, we could still guess the true values, especially for binary classification. For regression problems, even though labels can be encrypted with homomorphic encryption, it will be extremely time consuming for modeling. In practical tasks, there will be a trade-off between the security protection and the computational efficiency. 

\textit{\textbf{Features.}} 
On each client, local features were encoded before given to the master for global feature sampling. So the master will not know the real meaning of features. 

\textit{\textbf{Communication.}} Encryption methods such as RSA and AES can be applied to secure everything (model intermediate values, sample indices, e.t.c.) communicated during the training and prediction. 

\textit{\textbf{Model Storage.}}
The entire model was distributed across all clients. For each node, the client would store the corresponding split information only if the split feature is on local machine. If not, it only stored the structure of the current node. Clients knew nothing about each other including whose features were selected and at which tree nodes. Master can optionally keep a copy of the entire model.

\section{Experimental Studies}
\subsection{Experimental Setup}
In this section, we used 9 benchmark data sets, including one real-world data set \textit{target marketing} and 8 public data sets from UCI \citep{Dua:2019, sakar2019comparative, fernandes2015proactive, hamidieh2018data}, as shown in Table \ref{tab:datasets1}. Different sample sizes and feature spaces were considered, and the accuracy, efficiency and robustness of our proposed framework were tested for both classification and regression problems. In our experiments we did not pursue absolute accuracy and instead tested whether the performance of our methods is at the same level as the non-federated approach, i.e., lossless. The \textit{target marketing} data set was collected from two totally different domains. One of them was from an e-commerce company and contains 84 features, and the other one was from a bank which provided 11 features. Before modeling all the sensitive information was protected. Three main series of experiments were conducted in this section, including experiments with two data providers, experiments with multiple data providers, and analysis of prediction efficiency. The details of each test are given in the following subsections.

\subsection{Experiments with Two-Party Scenarios}

In this part, exposed UCI data sets were vertically and randomly separated by feature dimension and placed on two different client servers ($M=2$), each containing half of the feature space from original data. 
For \textit{target marketing}, it was also placed on two different client servers, of which each contained several business domains. 
The experiments in this section are summarized as following:

\begin{itemize}[leftmargin=*]
 	\item {\textbf{Federated Logistic/Linear Regression} (F-LR): We jointly trained logistic/linear regression models, where data is kept locally and the model is partly stored in each client.}

	\item{\textbf{Non-Federated Forest} (NonFF): All data were integrated together for Random Forest modeling.}
	\item{\textbf{Random Forest 1} (RF1): Partial data from the $1^{st}$ client was used to build a random forest model.} 
	\item{\textbf{Random Forest 2} (RF2): Partial data from the $2^{nd}$ client was used to build a random forest model.}
	
	\item{\textbf{Federated Forest} (FF): This is our proposed model, which two parties jointly learn a random forest. Data were kept locally and model was partly stored in each client.}
\end{itemize}







We conducted the experiments on both classification and regression problems, and present the results of accuracy and RMSE in Table \ref{tab:datasets1}. We found that the performance of RF1 and RF2 were obviously worse than the NonFF and FF. Both RF1 and RF2 can be considered as modeling with data from one business domain, and the insufficient feature space resulted in imperfect study of the global knowledge. We also found in most tests that the regression models didn't perform very well. For the test on \textit{target marketing}, since direct aggregation of data between two institutions was not allowed, we only ran tests for RF1, RF2, F-LR and FF. The results show that FF performs as expected and a better accuracy is achieved by building models on different domains. 

\begin{table}[]
\centering
\caption{Classification and regression experiments\label{tab:datasets1}}
\small
\begin{tabular}{ c c c c c c c c c }
\hline
 Classification &  RF1   &  RF2 &  F-LR  &  NonFF &  FF  & p-value\\ \hline
\texttt{target marketing}  & 0.870 & 0.848 & 0.862 & - &  $0.890\pm 0.014$ &  - \\ \hline
\texttt{ionosphere}& 0.864 & 0.828 & 0.873 & $0.908\pm 0.019$ & $0.896\pm0.030$ & $\textbf{0.211}$\\ \hline
\texttt{spambase} & 0.844 & 0.831 & 0.873 & $0.943\pm0.005$ & $0.928\pm0.020$ & \textbf{0.065} \\ \hline
\texttt{parkinson \citep{sakar2019comparative}} & 0.849 & 0.849 & 0.829 & $0.859\pm0.018$ & $0.857\pm0.013$ & $\textbf{0.744}$ \\ \hline
\texttt{kdd cup 99} & 0.974 & 0.965 & - & $0.995\pm0.001$ & $0.995\pm 0.009$ & $0.012$\\ \hline
\texttt{waveform} & 0.745 & 0.743 & - & $0.826\pm0.008$ & $0.822\pm0.012$  & 0.029 \\ \hline
\texttt{gene} & 0.975 & 0.975 & - & $0.988\pm0.005$ & $0.982\pm0.006$ & \textbf{0.229}\\ \hline\hline
Regression & RF1   &  RF2 &  F-LR  &  NonFF &  FF  & p-value\\ \hline
\texttt{year prediction} & 10.47 & 10.72 & 9.56 & $9.537\pm0.003$ & $9.555\pm0.061$ & \textbf{0.058} \\\hline
\texttt{Superconduct \cite{hamidieh2018data}}& 19.74 & 17.49 & 17.52 & $15.369\pm0.118$ & $15.411\pm0.16$3 & \textbf{0.186}\\ \hline
\end{tabular}
\end{table}

For most of the data sets, NonFF and FF outperformed the other methods. In our method, we were building each tree by processing globally on every regional domain, which was same to the tree built by aggregating raw data together. Z-Test was applied to verify the lossless of our method compared with NonFF, of which the null hypothesis is that the means from two populations are equal at a given level of significance. For each data set, 40 rounds of tests on the NonFF and FF were performed and the \textit{p-value} of each Z-Test is given in Table \ref{tab:datasets1}. If the $\textit{p-value} \ge 0.05$, the null hypothesis cannot be rejected at the 0.05 level and there is no significant difference between the outputs of NonFF and FF. If $0.01 \le \textit{p-value} < 0.05$, the null hypothesis cannot be rejected at the 0.01 level. And statistically, we consider there exists a slight but acceptable difference for this range of $\textit{p-value}$. The null hypothesis should be rejected if $\textit{p-value} < 0.01$ with a significant difference between the means. By examining the $\textit{p-value}$ of each data set, we can find that there are six of them proved to have no significant difference between the results of NonFF and FF, and for the rest data sets the differences are slight. No null hypotheses were rejected.

Overall, we can safely confirm that the Federated Forest is a lossless solution for both classification and regression problems, which achieves the same performance as the non-federated random forest.

\subsection{Experiments with Multi-Party Scenario}
\begin{wrapfigure}[8]{r}{0.5\textwidth}
\vspace{.4cm}
\small
    \includegraphics[width=\linewidth]{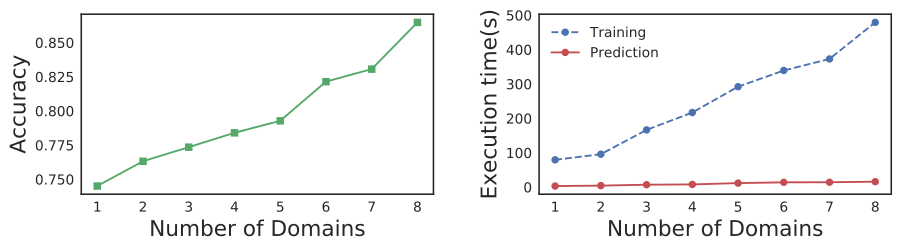}
	\caption{\small Accy. \& Exec. Time vs. \# of Domains}
	\label{fig:result of multi-servers}  
\end{wrapfigure}

In this part, we ran tests on the \textit{parkinson} data set to verify whether the Federated Forest is capable of conjoining more than two domains effectively and if a reasonable improvement on accuracy could be achieved. We chose \textit{parkinson} to run the test since it already contains eight clearly categorized sub-domains. As for tests of training and prediction efficiency, we duplicated data for ten times. In the tests, each time we added one domain into the federated model, and we recorded the accuracy, training and prediction time. As shown in Figure \ref{fig:result of multi-servers}, the accuracy of Federated Forest improved consistently. The training execution time was almost linearly with respect to to the number of domains, which is to be expected because all features are be examined in tree building. For the prediction time, though more domains and features were added, the difference in execution time was negligible. The results demonstrate that our new prediction algorithm is very effective when handling multiple regional domains.



\vspace{-.25cm}
\subsection{Prediction Efficiency}
\begin{wrapfigure}[9]{r}{0.5\textwidth}
\vspace{-.5cm}
\small
	\includegraphics[width=\linewidth]{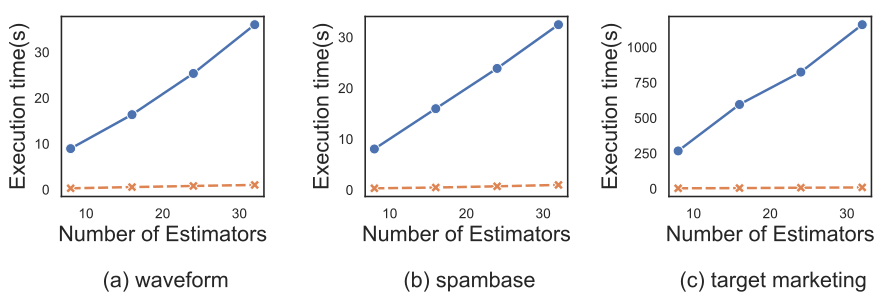}
	\caption{\small Prediction Time vs. Number of Estimators}
	\label{fig:estimator}
\end{wrapfigure}

\begin{wrapfigure}[9]{l}{0.5\textwidth}
\vspace{-.4cm}
\small
    \includegraphics[width=\linewidth]{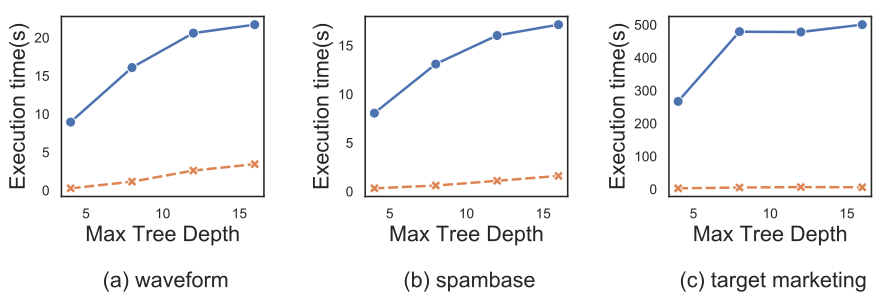}
	\caption{\small Prediction Time vs. Max Depth}
	\label{fig:depth}
\end{wrapfigure}

\begin{wrapfigure}[8]{r}{0.5\textwidth}
\vspace{-.5cm}
\small
    \includegraphics[width=\linewidth]{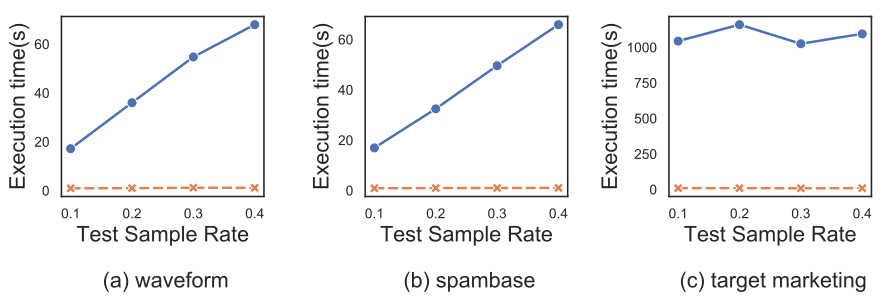}
	\caption{\small Prediction Time vs. Test Sample Size}
	\label{fig:test rate}
\end{wrapfigure}

In this part, we compared the efficiency of our new prediction method with the classical prediction approach. We used \textit{target marketing}, \textit{spambase} and \textit{waveform} data sets as the examples. We ran all the tests for 20 times and report the average results, as shown in Figures \ref{fig:estimator}, \ref{fig:depth} and \ref{fig:test rate}. The solid lines with dot marker represent the results of classical prediction method, and the dash lines with x marker represent our proposed prediction method.

Firstly, we set the maximum tree depth to 4 and changed the number of estimators from 8 to 32, and the results were shown in Figure \ref{fig:estimator}. It can be seen that our method produced a strong improvement on the prediction efficiency. Though the execution time of both methods increased linearly respect to the number of estimators, the slope varied dramatically between our method and the classical prediction method. For the classical method, there are multiple rounds of communication in each node during prediction. But in our method, there is only one round of communication for each tree.


Secondly, we set the number of estimators to 8, and adjusted the maximum tree depth from 4 to 16. As shown in Figure \ref{fig:depth}, our method outperformed the classical prediction method again. By increasing the maximum tree depth, the growth rate of prediction time for both methods gradually slowed down and stabilized. This is because by setting the maximum depth to a large number, the tree building may early stop due to pre-pruning and the actual tree depth will be smaller. In our method, no matter how deep the tree is or how many leaf nodes are created, communication was only executed once for each tree.

Finally, we fixed the number of estimators and maximum tree depth, and changed the test sample rate from 0.1 to 0.4, as shown in Figure \ref{fig:test rate}. Because the classical approach has a strong linear correlation with the sample size, we found that its results presented a linear growth trend. Meanwhile the execution time of our method changed very slowly, which shows our method is robust to prediction sample size. 


Overall, our new prediction method had been proved to be highly efficient.

\section{Conclusions}
In this paper, we proposed a novel tree-based machine learning model, called \textit{Federated Forest}, which is lossless with respect to the model accuracy and protects data privacy. A secure cross-regional machine learning system was developed based on it, which allows a learning model to be jointly trained across different clients with the same user samples but different attribute sets. The raw data on each client are not exposed and exchanged to other clients during the modeling. A novel prediction algorithm was proposed which could largely reduce the communication overhead and improve the prediction efficiency. Data privacy was secured by redesigning the tree algorithms, deploying encryption methods and establishing a third-party trusted server. Raw data will never be directly exchanged, only limited amount of intermediate values between each party. We performed experiments on both real-world and UCI data sets, showing the superior performance in classification and regression tasks, and the proposed Federated Forest was proven to be as accurate as the non-federated random forest that requires gathering the data into one place. The efficiency and robustness of our proposed system have also been verified. Overall, the Federated Forest overcomes the challenges of the \textit{data islands} problem and privacy protection in a brand new approach, and it can be deployed for real-world applications. 

\section*{Acknowledgement}
Special thanks to Chentian Jin for valuable discussions and feedback. 



\newpage
\section*{APPENDIX}
\subsection*{Reproducibility}
Our model is implemented with Python 3.6, Scikit-learn 0.20, Numpy 1.15.4, python-paillier 1.4.1 and mpi4py 3.0.0. We train/evaluate our model on servers each with 4 CPU cores and Centos 7.0. 
The information of all used data sets are given in Table \ref{datasets}. 
\begin{table}[!htbp]
\centering
\caption{Data sets \label{datasets}}
\begin{tabular}{cccc}
\hline
Classification                                  & Size   & Features  & Classes    \\ \hline
\texttt{target marketing}                       & 156198 & 95(11/84) & 2          \\ \hline
\texttt{ionosphere}                             & 351    & 34        & 2          \\ \hline
\texttt{spambase}                               & 4601   & 57        & 2          \\ \hline
\texttt{parkinson \citep{sakar2019comparative}} & 756    & 754       & 2          \\ \hline
\texttt{kddcup99}                               & 4M     & 42        & 23         \\ \hline
\texttt{waveform}                               & 5000   & 21        & 3          \\ \hline
\texttt{gene}                                   & 801    & 20531     & 5          \\ \hline\hline
Regression                                      & Size   & Features  & Range      \\ \hline
\texttt{year prediction}                        & 515345 & 90        & 1922-2011  \\ \hline
\texttt{Superconduct \cite{hamidieh2018data}}   & 21263  & 81        & 0.0002-185 \\ \hline
\end{tabular}
\end{table}

\subsection*{Pseudo-code for FF-Regressor\label{ref:regressor}}
The main difference between regression and classification problem lies in the generation of leaf node result and the final predictions. The following is the pseudo-code of regression problem, where the difference from the classification problem is in the line 7 of Algorithm \ref{alg-re-dt-client}, line 9 of Algorithm \ref{alg: re-dt-master} and line 5 of Algorithm \ref{alg: re-pred-master}.

\begin{algorithm}[!htbp]
	\SetAlgoLined
	\LinesNumbered 
    \SetKwInOut{Input}{Input}\SetKwInOut{Output}{Output}
    \SetKwFunction{Tree}{TreeBuild}%
	\Input{Data set $\mathcal{D}_{i}$ on client $i$;\\
	Local features $\mathcal F_{i} =\emptyset$ or $\mathcal F_{i} = \{f_{A}, f_{B}, \cdots \}$;\\
	Homomorphic encrypted label $y$;}
	\Output{Partial Federated Forest Model on Client $i$}
	\SetKwProg{TreeGenerateAlg}{Function}{}{end}
	\While{tree\_build is True}{
	Receive $\mathcal{F}_i^{'} \subset \mathcal{F}_i$ and $\mathcal{D}_i^{'} \subset \mathcal{D}_i$ for current tree building;\\
    \TreeGenerateAlg{\Tree($\mathcal{D}_{i}^{'}$, $\mathcal{F}_i^{'}$, $y$)}{
    Create empty tree node;\\
    \If{the pre-pruning condition is satisfied}{
    Mark current node as leaf node;\\
    Assign leaf label by averaging;\\ 
    \Return{leaf node};}
    $p, f^{*} \leftarrow -\infty, None$;\\
	\If{$\mathcal F_{i}^{'} \neq\emptyset$}{
	Compute impurity improvement $p$ for any $f \in \mathcal F_{i}^{'}$ and find local maximum $p_i$;\\
	Record local best split feature $f^*$ and split threshold;\\}
	Send encrypted $p_i$ to master;\\
	\eIf{receive the split message from master}{
		/* Global best split feature  $f_{global}^*$ is from itself */\\
		is\_selected $\leftarrow$ True; \\
		Split samples and send sample indices of left and right subtrees to master;\\}{
	Receive sample indices of left and right subtrees;\\}
    left\_subtree $\leftarrow$ \Tree($\mathcal{D}_{i\_left}^{'}$, $\mathcal{F}_i{'}$, $y_{left}$);\\
	right\_subtree $\leftarrow$ \Tree($\mathcal{D}_{i\_right}^{'}$, $\mathcal{F}_i^{'}$, $y_{right}$);\\
	\If{is\_selected is True}{
	Save $f^{*}$ and split threshold to tree node;}
	Save subtrees to tree node;\\
	\Return{tree node;}}
	Append current tree to forest;\\
	\Return{Partial Federated Forest Model on Client $i$;\\}}
	\caption{Federated Forest -- Client}
	\label{alg-re-dt-client}
\end{algorithm}

\begin{algorithm}[!htbp]
    \SetAlgoLined
	\LinesNumbered 
	\SetKwInOut{Input}{Input}\SetKwInOut{Output}{Output}
	\SetKwFunction{Tree}{TreeBuild}%
	\SetKwProg{ForestGenerateAlg}{Function}{}{end}
	\Input{Indices of $\mathcal{D}$; \\
		Encoded features $\mathcal F = \mathcal{F}_{1} \cup \mathcal{F}_{2} \cup \cdots \cup \mathcal{F}_{M}$; \\
		Encrypted label $y$;\\}
	\Output{Complete Federated Forest Model}
		/*Build trees for forest recurrently*/\\
		\While{tree\_build is True}{
		Broadcast randomly selected samples $\mathcal D^{'}$;\\
        Randomly select features $\mathcal{F}_i^{'}$ from $\mathcal{F}_i$ and send to client $i$;\\
		\ForestGenerateAlg{	\Tree($\mathcal D^{'}$, $\mathcal F^{'}$, $y$)}{
	        Create empty tree node;\\
			\If{the pre-pruning condition is satisfied}{Mark current node as leaf node;\\
				Assign leaf label by averaging;\\
				\Return{leaf node};}
				Receive encrypted $\{p\}_{i=1}^M$ and related information from all clients;\\
				Take $j = argmax(\{p\}_{i=1}^M)$ and notify client $j$;\\
				Receive split indices from client $j$ and broadcast;\\
				left\_subtree $\leftarrow$ \Tree($\mathcal D_{left}^{'}, \mathcal F^{'}, y_{left}$);\\
				right\_subtree $\leftarrow$ \Tree($\mathcal D_{right}^{'}, \mathcal F^{'}, y_{right}$);\\
				Save subtrees and split info to tree node;\\
				\Return{tree node;}
			}
			Append current tree to forest;\\
			\Return{Complete Federated Forest Model;}
		}
	\caption{Federated Forest -- Master}
	\label{alg: re-dt-master}
\end{algorithm}

\begin{algorithm}[!htbp]
	\SetAlgoLined
	\LinesNumbered 
	\SetKwInOut{Input}{Input}\SetKwInOut{Output}{Output}
	\Input{Partial federated forest model saved on $i$th client;\\
		Encoded features $\mathcal F_{i}$ on $i$th client;\\
		Test set $\mathcal D^{test}_{i}$ on $i$th client;}
	\Output{Samples IDs $S_i^l$ of leaf $l$ on $T_i$, $l \in \mathcal{L}$}
	\SetKwProg{TreePredictAlg}{Function}{}{end}
	\SetKwFunction{Tree}{TreePredict}%
	\While{TreePrediction is True}{
	\TreePredictAlg{\Tree(\small{$T_{i}$, $\mathcal D^{test}_{i}$, $\mathcal F_{i}$})}{
		\eIf{is\_leaf is True}{Return sample IDs $S_{i}^{l}$ and leaf label;}{
			\eIf{$T_{i}$ keeps the split info of current node}
			{	Split samples into subtrees based on threshold;\\
				left\_subtree $\leftarrow$ \Tree({$T_{i\_left}$, $\mathcal F_{i}$, $\mathcal D_{i\_left}^{test}$});\\
				right\_subtree $\leftarrow$ \Tree({$T_{i\_right}$, $\mathcal F_{i}$, $\mathcal D_{i\_right}^{test}$});\\
			}
			{
				left\_subtree $\leftarrow$ \Tree({$T_{i\_left}$, $\mathcal F_{i}$, $\mathcal D_{i}^{test}$});\\
				right\_subtree $\leftarrow$ \Tree({$T_{i\_right}$, $\mathcal F_{i}$, $\mathcal D_{i}^{test}$});\\
			}
			Return left and right subtrees;\\
		}	
		Send $S_{i} = \{S_{i}^{1}, S_{i}^{2}, \cdots ,S_{i}^{l}, \cdots \}$ to master;
	}
	\Return{};}
	\caption{Federated Forest Prediction -- Client}
	\label{alg: re-pred-client}
\end{algorithm}

\begin{algorithm}[!htbp]
	\SetAlgoLined
	\LinesNumbered 
	\SetKwInOut{Input}{Input}\SetKwInOut{Output}{Output}
	\Input{Sample IDs $S$ of test set $\mathcal{D}^{test}$;\\}
	\Output{ Prediction of Random Forest}
	\While{TreePrediction is True}{
		Gather $\{S_{1}, S_{2}, \cdots ,S_{i}, \cdots \}$;\\
		Obtain $ \{ S^{1}, S^{2}, \cdots , S^{l}, \cdots \} $, where $S^{l} = S^{l}_{1} \cap S^{l}_{2} \cap \cdots \cap S^{l}_{M}$;\\
		Return label of leaf $l$ for samples in $S^l$, $l \in \mathcal{L}$;
	}
    Calculate forest predictions by averaging the results of trees;\\
    \Return{Final Predictions;}
	\caption{Federated Forest Prediction -- Master}
	\label{alg: re-pred-master}
\end{algorithm}

\subsection*{Notations In Proof}

\begin{itemize}[leftmargin=*]
	
	\item Sample IDs are denoted as $S$, and $S^{l}_{i}$ contains the sample IDs which fall into leaf $l$ of tree $T_{i}$. $S^{l}$ denotes the sample set of leaf node $l$ in the complete binary tree model $T$.
	
	\item The test sample set is $ \mathcal H$, and the single sample is $h \in  \mathcal H$.
	
	\item $W_{i}$ is the set of decision making paths of sample $h$ that goes through the binary tree to fall into the leaf node of $T_{i}$. For the tree $T_i$, it is possible that $h$ falls into more than one leaf, due to our model storage strategy. 
	
	\item $w^*$ is the decision making path of the sample $h$ that goes through the complete binary tree to fall into the leaf node in $T$. For the complete tree $T$, if sample $h$ fall into one leaf, then it cannot fall into another leaf. It means that any leaf $l$ and $g$ in $T$, $S^l \cap S^g = \emptyset$.
	
	\item The complete tree $T$ on master is defined as
	$T = T_{1} \cup T_{2} \cup \cdots \cup T_{M}$.
	\item Detailed descriptions of notations are shown in Table \ref{Notations1}.
	
\end{itemize}

\begin{table}[!htbp]
	\begin{center}
    \caption{Notations}
	\label{Notations1}
 	\begin{tabular}{c l}\hline
		Notation & Description \\\hline
 		$M$ & number of regional domains \\\hline
		$\mathcal{D}_{i}$ & data set held by client $i$ \\\hline
		$N$ & total number of samples in training \\\hline
		$\mathcal{D}$ & entire data set $\mathcal{D} = \{\mathcal{D}_{1}, \mathcal{D}_{2}, \cdots , \mathcal{D}_{M}\}$ \\\hline 	
		$\mathcal{F}_{i}$ & feature space of $\mathcal{D}_{i}$ \\\hline
		$\mathcal{F}$ & entire feature space of $\mathcal{D}$, $\mathcal{F} = \mathcal{F}_{1} \cup \mathcal{F}_{2} \cup \cdots \cup \mathcal{F}_{M}$\\\hline 
		$y$ & labels \\\hline 
		$T_i$ & partial decision/regression tree stored on $i$th client \\\hline
		$T$ & complete tree $T = T_{1} \cup T_{2} \cup \cdots \cup T_{M}$
		 \\\hline 	
		$\mathcal{L}$ & leaf nodes set of the entire tree\\\hline 
		$l, g$ & leaf node of the current tree, $l, g \in \mathcal{L}$ \\\hline
		$O$ & lowest common ancestor of $l, g$ in $T$ \\\hline
		$S$ & the sample IDs of entire data set $\mathcal{D}$\\\hline		
		$S^{l}_{i}$ & the sample IDs which fall into leaf $l$ of tree $T_{i}$\\\hline
		$S^{l}$ & the sample IDs which fall into leaf $l$ of complete tree $T$ \\\hline
		$h$ & single test sample \\\hline		
		$\mathcal{H}$ & entire test sample set\\\hline
		$W_{i}$ & the set of decision making paths of sample $h$ on $T_i$\\\hline
		$w^*$ & decision making path of sample $h$ on $T$\\\hline
		$k$ & maxmium tree depth\\\hline

	\end{tabular} 
	\end{center} 
\end{table}

\subsection*{Proof of the Proposition 1\label{Proof}}
For the prediction process, samples $S$ will go through the client tree $T_i$ and fall into one or multiple leaves. For any leaf $l$ of the complete tree $T$, the sample IDs $S^{l}$ in leaf $l$ can be obtained by taking intersection of $\{S^{l}_{i}\}_{i=1}^{M}$, that $S^{l} = S^{l}_{1} \cap S^{l}_{2} \cap \cdots \cap S^{l}_{M}$.

\begin{proof}
In order to prove $S^{l} = S^{l}_{1} \cap S^{l}_{2} \cap \cdots \cap S^{l}_{M}$, we will prove:
\begin{itemize}
    \item $S^{l} \subseteq S^{l}_{1} \cap S^{l}_{2} \cap \cdots \cap S^{l}_{M}$
    \item $S^{l} \supseteq S^{l}_{1} \cap S^{l}_{2} \cap \cdots \cap S^{l}_{M}$
\end{itemize} 

\textit{Proof of} $S^{l} \subseteq S^{l}_{1} \cap S^{l}_{2} \cap \cdots \cap S^{l}_{M}$:

For any sample $h$ in the leaf $l$ of the complete tree $T$, $h \in S^l$. $w^*$ denotes its decision making path from root to leaf node.
For model $T_i$ on each client $i$, if the model stores split information at the current node, it is determined according to the threshold whether this sample enters the left or right subtree. If the current model does not store split information at this node, the sample enters left and right subtrees simultaneously. Therefore for sample $h$, its decision making path $w^*$ on the complete tree $T$ must be subset of its decision making path $W_i$ on any client $i$. Then we have $w^* \subseteq W_i, 1 \leq i \leq M$, which is equivalent to $h \in S_i^l, 1\leq i\leq M$. 
Because of this we can safely say that $h \in S^{l}_{1} \cap S^{l}_{2} \cap \cdots \cap S^{l}_{M}$ for any $h$ in $S^l$. Then we can prove that $S^{l} \subseteq S^{l}_{1} \cap S^{l}_{2} \cap \cdots \cap S^{l}_{M}$.

\textit{Proof of} $S^{l} \supseteq S^{l}_{1} \cap S^{l}_{2} \cap \cdots \cap S^{l}_{M}$:

    Assume that sample $h$ doesn't belong to leaf node $l$ but belongs to $g$ in complete model $T$, which is $h \notin S^{l}$ and $ h \in S^{g}$. Besides, we assume $h \in S^{l}_{1} \cap S^{l}_{2} \cap \cdots \cap S^{l}_{M}$.
    
    $\Longrightarrow h \in S^{g}_{1} \cap S^{g}_{2} \cap \cdots \cap S^{g}_{M}$, obtained by the above proof.
    
    $\Longrightarrow h \in (S^{g}_{1} \cap S^{g}_{2} \cap \cdots \cap S^{g}_{M}) \cap ( S^{l}_{1} \cap S^{l}_{2} \cap \cdots \cap S^{l}_{M})$
    
    $\Longrightarrow h \in (S^{g}_{1} \cap S^{l}_{1}) \cap (S^{g}_{2} \cap S^{l}_{2}) \cap \cdots \cap (S^{g}_{M} \cap S^{l}_{M})$
    
    That is to say, sample $h$ will fall into the leaf node $g$ and $l$ at the same time in every model stored on client.
    
    $ \because $ In the same binary tree structure, the path from a child node to the root node is fixed  and unique.
    
    Under the complete tree structure, the path set of the leaf node $g$ and $l$ up to the root node is $w^l \cup w^j$. And the lowest common ancestor node exists and is uniquely set to $O$.
    
    So  $(w^l \cup w^j) \subseteq  W_{i} \Longrightarrow  (w^l \cup w^j) \in (W_{1} \cap W_{2} \cap \cdots \cap W_{M})$
    
    So no platform stores the information of the node $O$.
    
    $\Longrightarrow T \neq T_{1} \cup T_{2} \cup \cdots \cup T_{M}$
    
    This contradicts to $T = T_{1} \cup T_{2} \cup \cdots \cup T_{M}$.
    
    Therefor the hypothesis doesn't hold.
    
    $\Longrightarrow h \notin S^{l} \Longrightarrow h \notin S^{l}_{1} \cap S^{l}_{2} \cap \cdots \cap S^{l}_{M}$
    
    $\Longrightarrow S^{l} \supseteq S^{l}_{1} \cap S^{l}_{2} \cap \cdots \cap S^{l}_{M}$

In summary, we can prove $S^{l} = S^{l}_{1} \cap S^{l}_{2} \cap \cdots \cap S^{l}_{M}$.
    
\end{proof}

\subsection*{Communication Complexity Analysis\label{comm_comp}}
Here we give a brief analysis on communication complexity. There are mainly three types of communication during the training, where $M$ is the number of regional domains:
\begin{itemize}[leftmargin=*]
\item\textit{\textbf{Send and receive.}}
Master sends randomly selected features to each client in every turn for tree building and the client who saves the global optimal feature sends the sample split indices of this feature to master when building the node. The communication complexity is $O(1)$.

\item\textit{\textbf{Broadcast.}}
Master broadcasts sample indices for each tree node construction. The communication complexity is $O(M)$.

\item\textit{\textbf{Gather.}}
Master gathers and compares the impurity improvement of features at every turn for node building. It also gathers sample sets of all leaves on each tree stored by clients in the prediction process. The communication complexity is $O(M)$.
\end{itemize}

Since the maximum depth is $k$, in a tree, there are at most $2^{k-1}-1$ intermediate nodes and $2^{k-1}$ leaf nodes. Take the process of building a tree for example, the communication complexity of the whole system in training phase is $O(2^{k}(M+1))$. For the prediction phase, if not optimized, the communication complexity is $O(2^{k-1}M)$, otherwise, the optimized communication complexity is $O(M)$.



\end{document}